\documentclass[lettersize,journal]{IEEEtran}
\usepackage{amsmath,amsfonts}
\usepackage{algorithmic}
\usepackage{algorithm}
\usepackage{array}
\usepackage[caption=false,font=normalsize,labelfont=sf,textfont=sf]{subfig}
\usepackage{textcomp}
\usepackage{stfloats}
\usepackage{url}
\usepackage{verbatim}
\usepackage{graphicx}
\usepackage{cite}
\usepackage{booktabs}
\usepackage{color}
\usepackage{multirow}

\begin{document}
\title{Progressive Feedback-Enhanced Transformer for Image Forgery Localization}
\author{Haochen Zhu, Gang Cao,~\IEEEmembership{Member,~IEEE}, Xianglin Huang
\thanks{Haochen Zhu, Gang Cao and Xianglin Huang are with the State Key Laboratory of Media Convergence and Communication, Communication University of China, Beijing 100024, China, and also with the School of Computer and Cyber Sciences, Communication University of China, Beijing 100024, China (e-mail: \{zhuhaochen, gangcao, huangxl\}@cuc.edu.cn).}}

\maketitle

\begin{abstract}
Blind detection of the forged regions in digital images is an effective authentication means to counter the malicious use of local image editing techniques. Existing encoder-decoder forensic networks overlook the fact that detecting complex and subtle tampered regions typically requires more feedback information. In this paper, we propose a Progressive FeedbACk-enhanced Transformer (ProFact) network to achieve coarse-to-fine image forgery localization. Specifically, the coarse localization map generated by an initial branch network is adaptively fed back to the early transformer encoder layers, which can enhance the representation of positive features while suppressing interference factors. The cascaded transformer network, combined with a contextual spatial pyramid module, is designed to refine discriminative forensic features for improving the forgery localization accuracy and reliability. Furthermore, we present an effective strategy to automatically generate large-scale forged image samples close to real-world forensic scenarios, especially in realistic and coherent processing. Leveraging on such samples, a progressive and cost-effective two-stage training protocol is applied to the ProFact network. The extensive experimental results on nine public forensic datasets show that our proposed localizer greatly outperforms the state-of-the-art on the generalization ability and robustness of image forgery localization. Code will be publicly available at \url{https://github.com/multimediaFor/ProFact}. 
\end{abstract}


\section{Introduction}
\IEEEPARstart{T}{he} rapid development of digital image processing techniques and tools enables people to manipulate image content easily, resulting in widespread forged images. The credibility of digital images has been significantly diminished, since the forged images may be used for malicious purposes, such as fake news, academic fraud and criminal activities. It incurs serious threats to the multimedia authenticity in various applications. Moreover, such forged images are often indistinguishable to the naked eye. Therefore, it is necessary to develop forensic algorithms for identifying the forged regions within images in order to prevent malicious image forgeries.

In recent years, many image forgery localization methods have been proposed to expose the local manipulations that alter the semantic content of an image, such as splicing, copy-move and inpainting \cite{verdoliva2019multimedia}. The post-processing including blurring, compression, contrast adjustment, etc., is frequently involved in concealing the visual clues of forged regions. A fundamental assumption in image forgery localization is that there exists some inconsistency between the forged and authentic regions within an image. Such inconsistency is often hidden within subtle details or signals. Early methods reveal forgeries by analyzing the abnormality of handcrafted features, such as lens distortion \cite{mayer2018accurate}, color filter array (CFA) artifacts \cite{popescu2005exposing}, and noise \cite{cozzolino2015splicebuster}. Such methods are typically limited to targeted types of operations due to the reliance on prior statistical models. Recently, many deep learning-based methods \cite{wu2019mantra,dong2022mvss,zhou2018learning,kwon2022learning,salloum2018image,bappy2019hybrid,liu2022pscc,bayar2018constrained,huh2018fighting,mayer2019forensic,cozzolino2019noiseprint,zhuang2021image,wang2022objectformer,wu2022robust,hu2020span,guillaro2023trufor,li2023edge,guo2023hierarchical,xia2024mmnet,liu2023fedforgery,liu2023hierarchical} have been proposed for image forgery localization. With powerful feature representation capability and abundant training samples, such deep learning-based methods significantly outperform the traditional ones in terms of localization performance.

\begin{table*}[!tb]
\caption{Brief summary of image forgery localization methods based on deep learning. ‘(-)’ denotes unavailable numbers in the literature.} 
\label{tab:table1}
\footnotesize
\centering
\begin{tabular}{llll}
\toprule 
        Method & Backbone & Feature & Training set \\ \midrule
        MFCN, 2018\cite{salloum2018image}  & VGG & surface and edge artifacts & CASIAv2 (5,123) \\ \midrule
        RGB-N, 2018\cite{zhou2018learning}  & Faster RCNN & RGB + noise  & simply synthetic images (42K) \\ \midrule
        EXIF-SC, 2018\cite{huh2018fighting}  & Siamese Resnet & EXIF  inconsistency & pristine images (400K) \\ \midrule
        ForSim, 2019\cite{mayer2019forensic}  & Siamese MISLnet & forensic similarity & pristine images from 95 cameras (47,785) \\ \midrule
        Noiseprint, 2019\cite{cozzolino2019noiseprint}  & Siamese DnCNN & camera model fingerprint & 4 pristine datasets with 125 cameras \\  \midrule
        ManTra, 2019\cite{wu2019mantra}  & Wider VGG & anomalous features & simply synthetic images (102K) \\  \midrule
        H-LSTM, 2019\cite{bappy2019hybrid}  & Patch-LSTM & RGB + resampling  & simply synthetic images (65K) \\  \midrule
        SPAN, 2020\cite{hu2020span}  & Wider VGG & local self-attention & simply synthetic images (102K) \\ \midrule
        DFCN, 2021\cite{zhuang2021image}  & Dense FCN & RGB & synthetic images by Photoshop script (-) \\  \midrule
        TransForensics, 2021\cite{hao2021transforensics}   & CNN + Transformer & dense self-attention  & CASIAv2, Coverage and IMD \\  \midrule
        MVSS, 2022\cite{dong2022mvss}  & FCN & noise + edge artifacts& CASIAv2 (12,554) or DEFACTO (84K) \\  \midrule
        OSN, 2022\cite{wu2022robust}  & SE-U-Net & RGB & simply synthetic images (15K) \\ \midrule
        PSCC, 2022\cite{liu2022pscc}  & HRNet & spatio-channel correlation & simply synthetic images (380K) \\ \midrule   
        ObjectFormer, 2022\cite{wang2022objectformer}   & CNN + Transformer & RGB + high-frequency & simply synthetic images (62K) \\  \midrule
        \multirow{2}*{CAT, 2022\cite{kwon2022learning}}  & \multirow{2}*{HRNet} & \multirow{2}*{RGB + compression artifacts} & simply synthetic images (824K) + \\
        & & & 3 standard datasets (51K) \\ \midrule
        MSMG, 2022\cite{wang2022msmg}   & CNN + Transformer & RGB + edge artifacts  & CASIAv2, Coverage, NIST16 and IMD \\  \midrule
        \multirow{2}*{EMT, 2023 \cite{lin2023image}}   & \multirow{2}*{CNN + Transformer} & \multirow{2}*{RGB + noise + edge artifacts} & CASIAv2, NIST16, Coverage, CoMoFoD and \\
        & & & DEFACTO (112K)\\  \midrule  
        ERMPC, 2023\cite{li2023edge}  & Resnet & RGB + noise & simply synthetic images (60K) \\  \midrule
        \multirow{2}*{{HiFi, 2023\cite{guo2023hierarchical}} } & \multirow{2}*{HRNet} & \multirow{2}*{RGB + frequency} & simply synthetic images (300K) 
 + \\
        & & & CNN-synthesized + pristine images (1,410K) \\ \midrule
        \multirow{2}*{{TruFor, 2023\cite{guillaro2023trufor}}}  & \multirow{2}*{Segformer} & \multirow{2}*{RGB + noiseprint} & simply synthetic images (824K) + \\
        & & & 3 standard datasets (51K) \\ \midrule
        TBFormer, 2023\cite{liu2023tbformer}  & Resnet & RGB + noise & simply synthetic images (150K) \\  \midrule
        IML-ViT, 2023\cite{ma2023iml}  & Vision Transformer & RGB + edge artifacts & CASIAv2 (12,554) \\  \midrule
        ProFact (ours) & Segformer & feedback-enhanced & realistic synthetic images by MBH (55K) \\ \bottomrule
\end{tabular}
\end{table*}

In order to uncover the subtle tampering traces, most methods employ convolutional neural networks (CNNs) with an encoder-decoder structure to capture distinctive patterns in the forged image, such as noise \cite{wu2019mantra,dong2022mvss,zhou2018learning,guillaro2023trufor,liu2023hierarchical}, compression \cite{kwon2022learning} and edge artifacts \cite{dong2022mvss, salloum2018image,li2023edge}. Simultaneously, the ground truth supervises the final output to guide the network to learn effective feature representations for identifying forged regions. However, such straight forensic networks solely rely on the direct supervision of final outputs. They fail to fully harness effective feature representations at multiple encoding stages and lack feedback and refinement of intermediate results.

To this end, here we propose a progressive feedback-enhanced transformer (ProFact) network for image forgery localization. The two cascaded branches with similar architecture are connected through a coarse-to-fine mechanism. Guided by the coarse localization map generated from the initial branch, a feedback-enhanced branch is deployed to refine and correct the forensic features of intermediate encoder blocks for improving localization performance. Considering the local information and the size diversity of forged regions, we design a contextual spatial pyramid module to strengthen the local artifact features and adapt to forgery localization tasks at different scales.

Furthermore, the deep learning-based localizers heavily rely on abundant forged image samples with pixel-level labels, which poses a challenging requirement. Existing methods \cite{zhou2018learning,kwon2022learning,bappy2019hybrid,liu2022pscc} often simulate forged images through simple splicing or copy-move. However, directly pasting the segmented objects onto backgrounds would result in apparently unrealistic regions and boundaries. To improve the realism of synthetic samples, an effective generation method is proposed based on the combined use of digital matting, blending, and harmonization (MBH). It provides new insights for automatically generating large-scale forged images, which better align with real-world applications. In addition, the adopted progressive two-stage training strategy enables the ProFact to gradually learn better features, and thus further improve the localization capability. In the latter stage, the network is trained on the hard samples that are more challenging to detect at different scales.

Our main contributions are summarized as follows:
\begin{itemize}
\item We propose ProFact, a novel scheme for image forgery localization based on a feedback-enhanced transformer network. A coarse localization result is fed back to the early encoding layer and further refined by another encoder-decoder branch for generating a more accurate localization map.
\item We implement an effective strategy to automatically generate large-scale forged image samples for training the ProFact. It provides a new insight for achieving more realistic visual simulations of real-world forged images. 
\item Extensive experimental results verify that our proposed image forgery localizer achieves superior performance compared with the state-of-the-art on multiple benchmarks. 
\end{itemize}

The rest of this paper is organized as follows. Section II reviews previous related works. The proposed ProFact network and the training data generation strategy are elaborated in Section III. The experimental results and discussions are depicted in Section IV, followed by the conclusion drawn in Section V.

\section{RELATED WORK}
\subsection{Image Forgery Localization}

Forged image regions are generally difficult to locate due to the subtle forensic traces. Although many traditional methods \cite{mayer2018accurate,popescu2005exposing,cozzolino2015splicebuster} make efforts to capture the statistical and visual anomalies incurred by tampering operations, they are limited to forgery types and post-processing. To overcome such difficulty, plenty of deep learning-based forgery localizers have been proposed in recent years, which directly learn and optimize the feature representations towards forensics. Table \ref{tab:table1} summarizes the network and training deployments of such methods, and explains the novelty of our approach accordingly. In \cite{huh2018fighting,mayer2019forensic,cozzolino2019noiseprint}, Siamese network is used to learn the block-wise feature representations whose inconsistency within an image is treated as a forensic clue. For example, it involves the features from EXIF metadata \cite{huh2018fighting} and the processing histories inside and outside of a camera \cite{mayer2019forensic,cozzolino2019noiseprint}. 

The localizers based on block-wise inconsistency often suffer a high computational cost in testing phrases due to processing massive patch-level features. To attenuate such deficiency, many segmentation-based forgery localization methods have been proposed and achieve higher forensic performance. The forgery localization is treated as a supervised segmentation task in a straight encoder-decoder framework. The employed backbones include CNN\cite{zhou2018learning,bayar2018constrained,li2023edge}, long short-term memory (LSTM) network \cite{bappy2019hybrid}, and fully convolutional network (FCN) \cite{wu2019mantra,zhuang2021image,dong2022mvss,kwon2022learning,salloum2018image,guo2023hierarchical}. Specifically, ManTra \cite{wu2019mantra} formulates the forgery localization problem as a local anomaly detection task and contributes a LSTM-based solution. OSN \cite{wu2022robust} improves the robustness of localizers by modeling the noise incurred by online social networks in a U-Net framework. Liu et al.\cite{liu2023hierarchical} propose a novel hierarchical forgery classifier, which enhances forgery authentication in multi-channel scenarios by effectively learning robust mixed-domain representations. Furthermore, recent methods\cite{wang2022objectformer,wang2022msmg,lin2023image,guillaro2023trufor,liu2023tbformer,ma2023iml} leverage Transformer structures to better capture global dependencies and the feature difference between forged and authentic regions. Different from the single-stage or parallel two-stream framework used in the previous works, our proposed scheme adopts a cascaded dual-branch network to refine the forgery localization results. 

\begin{figure*}[!tb]
\centering
\includegraphics[width=\linewidth]{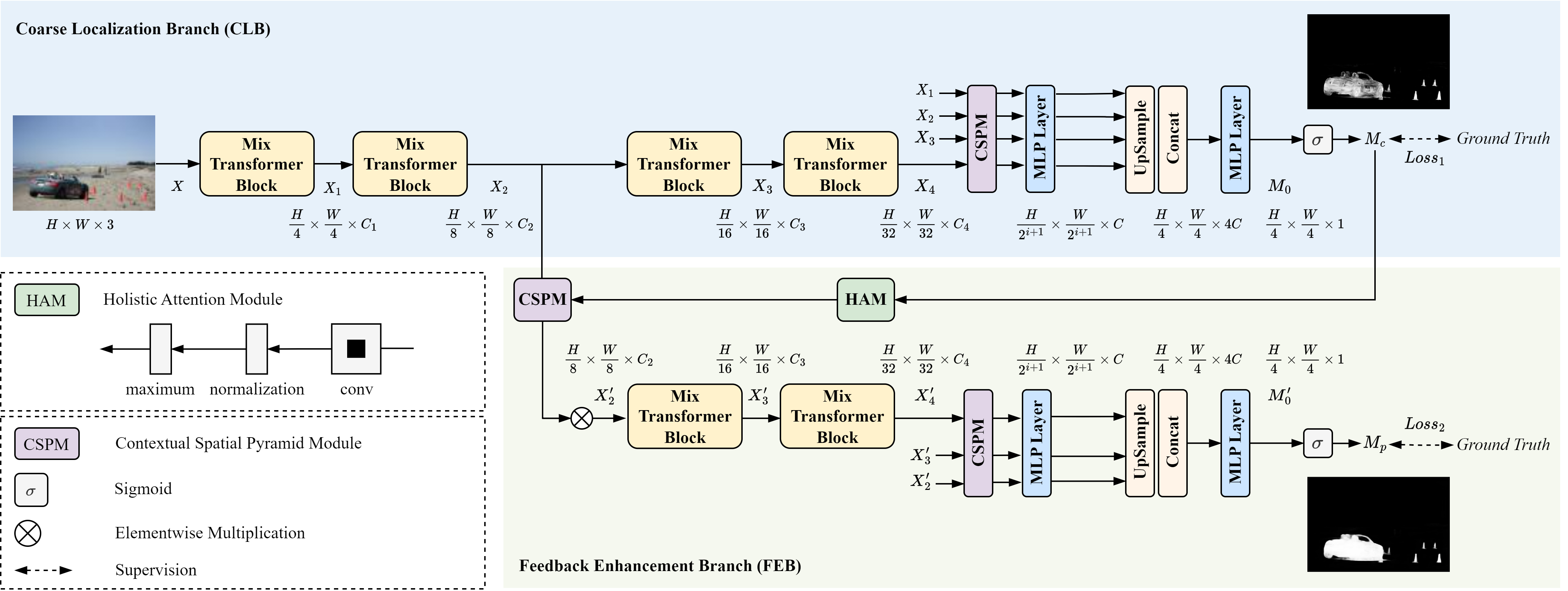}
% \captionsetup{skip=0pt}
\vspace{-0.5cm}
\caption{Proposed image forgery localization network ProFact. The coarse localization branch (top) generates a coarse localization map $M_c$ of the input forged image $X$. The feedback enhancement branch (down) predicts the final refined localization map $M_p$ by re-encoding the middle-level features of CLB with the attentive feedback.} \label{fig_1}
\end{figure*}

\subsection{Coarse-to-fine Mechanism}
Coarse-to-fine is an effective processing method used in various vision applications, such as segmentation \cite{pinheiro2016learning,zhang2019canet}, object localization \cite{song2020progressive,wei2020f3net,wu2019cascaded}, and denoising \cite{li2019progressive}. Such a strategy has also been adopted by a few forensic detectors. For the PSCC detector \cite{liu2022pscc} a densely connected pyramid structure is presented to supervise the masks, which are refined progressively from small to large scales. Zhuo et al.  \cite{zhuo2022self} create a coarse-to-fine network, where the separable refined subnetwork directly receives the naive feature map output by the coarse one. Different from such previous works and inspired by \cite{wu2019cascaded}, we design a cascaded transformer network to provide feedback from the coarse result to the intermediate encoding layers for achieving refined localization. Besides, we further adopt a progressive training strategy to optimize and improve the localization performance.

\subsection{Training Samples in Forgery Localization}
To mitigate the issue of limited training sample sets, data augmentation is widely used to increase the quantity of samples \cite{dong2022mvss,zhou2018learning}. For example, MVSS \cite{dong2022mvss} considers the flipping, blurring, compression, and adding or removing squared areas. However, the additional samples obtained through such data augmentation remain constrained. Zhuang et al. \cite{zhuang2021image} use Photoshop scripts to yield sufficient forged samples for training a localizer. Most state-of-the-art methods  \cite{wu2019mantra,wu2022robust,zhou2018learning,kwon2022learning,bappy2019hybrid,liu2022pscc} instead generate large-scale simple synthetic samples from the COCO dataset \cite{lin2014microsoft} with rough annotations, as shown in Table \ref{tab:table1}. Such images are created automatically by splicing any objects into another image at random positions, followed by locally random rotations and scaling. However, many unnatural artifacts are incurred around the splicing boundaries due to the unreasonable synthetic operations, resulting in unrealistic forged images. Such low-quality training samples would lead to a gap in simulating the real-world distribution, thereby degrading the testing performance of localizers.

\section{PROPOSED SCHEME}
In this section, the architecture design, training dataset generation and training strategy of our proposed image forgery localization network, i.e., ProFact, are presented in detail.
\subsection{Network Architecture}
\subsubsection{Overview}
The overall framework of our proposed forgery localization network is illustrated in Fig. \ref{fig_1}. Note that ProFact is a progressive refined network comprising two branches with similar architectures, named coarse localization branch (CLB) and feedback enhancement branch (FEB). Given a RGB image $X$ with size $H\times$$W\times$3, a coarse pixel-level tampering probability map ${M_c} \in {\left[ {0, 1} \right]^{H \times W}}$. ${M_c}$ is first generated by the CLB, which consists of four encoder stages and a contextual spatial pyramid module (CSPM) followed by a multilayer perceptron (MLP) decoder. To enhance representation ability of the encoder, the coarse map ${M_c}$ is backward fused to the intermediate transformer encoding feature ${X_2}$ via a holistic attention module (HAM) \cite{wu2019cascaded}. The resulting fused feature ${X'_2}$ is transformed to the final refined localization map ${M_p}$ by the FEB. Finally, the two branches are trained jointly under the supervision of forgery region ground truth.  

\subsubsection{Coarse Localization Branch}
We adopt Segformer \cite{xie2021segformer} as the backbone, which is a hierarchical network based on stacked Mix Transformer (MiT) encoding blocks. MiT involves common channel operations with self-attention mechanism. In contrast to the CNN used in most prior works, such a transformer backbone is better at capturing the long-range global dependency between different image regions \cite{dosovitskiy2020image}. To enhance the representative learning for tampering traces, the stacked MiT blocks followed by an efficient MLP decoder are adopted in the CLB. Multi-scale features bring sufficient spatial and semantic information for subsequent fusion and decoding. Since the backbone does not rely on positional encoding, our ProFact scheme is well-suited for forgery  localization tasks where the resolutions of test and training images are often different \cite{xie2021segformer}.

Given an input image $X$ of size $H\times$$W\times$3, it is first divided into non-overlapping $4\times4$ patches. Such patches are input to four cascaded MiT blocks for yielding multi-level features $\left \{ X_{1} ,X_{2},X_{3},X_{4} \right \}$, which enjoy the sizes 1/4, 1/8, 1/16, and 1/32 of $H\times$$W$, respectively. ProFact can effectively capture the high-resolution coarse-grained and low-resolution fine-grained forensic clues, and learn the context correlations between different regions. The multi-scale features ${X_i}$ are passed through the CSPM and fed into the lightweight MLP layers. The resulting feature map $M_{0} \in 
\mathbb{R} ^{1 \times H/4\times W/4 }$  is formulated as
\begin{equation}
\label{eq1}
{M_0} = {D_1}\left( {{X_1}, {X_2}, {X_3}, {X_4}} \right)
\end{equation}
where $D_1$ denotes the decoder of the coarse localization branch, which encompasses MLP layers, upsampling, and fusion processes. Specifically, the enhanced features ${\rm{CSPM}}\left( {{X_i}} \right)$ are unified with the same channel dimension $C$ by a linear layer, and then upsampled to ${\bar X_i}  \in 
\mathbb{R} ^{C \times H/4\times W/4 } $ . That is,
\begin{equation}
\label{eq2}
{\bar X_i} = {\rm{Upsample}}\left( {{\rm{Linear}}\left( {{\rm{CSPM}}\left( {{X_i}} \right)} \right)} \right)
\end{equation}
where $i = 1, 2, 3, 4$. Subsequently, ${\bar X_i}$  are concatenated and enforced by two consecutive linear layers. The yielded ${M_0}$ is converted to the coarse localization map  ${M_c}$ by the parameter-free bilinear upsampling followed by an element-wise sigmoid function $\sigma$,  as formulated in Eq.(\ref{eq3}) .
\begin{equation}
\label{eq3}
\begin{aligned}
&{M_0} = {\rm{Linear}}\left( {{\rm{Concat}}\left[ {{{\bar X}_1},{{\bar X}_2},{{\bar X}_3},{{\bar X}_4}} \right]} \right) \\
&{M_c} = \sigma \left( {{\rm{Upsample}}\left( {{M_0}} \right)} \right)
\end{aligned}
\end{equation}

\begin{figure}[!tb]
\centering
\includegraphics[width=\linewidth]{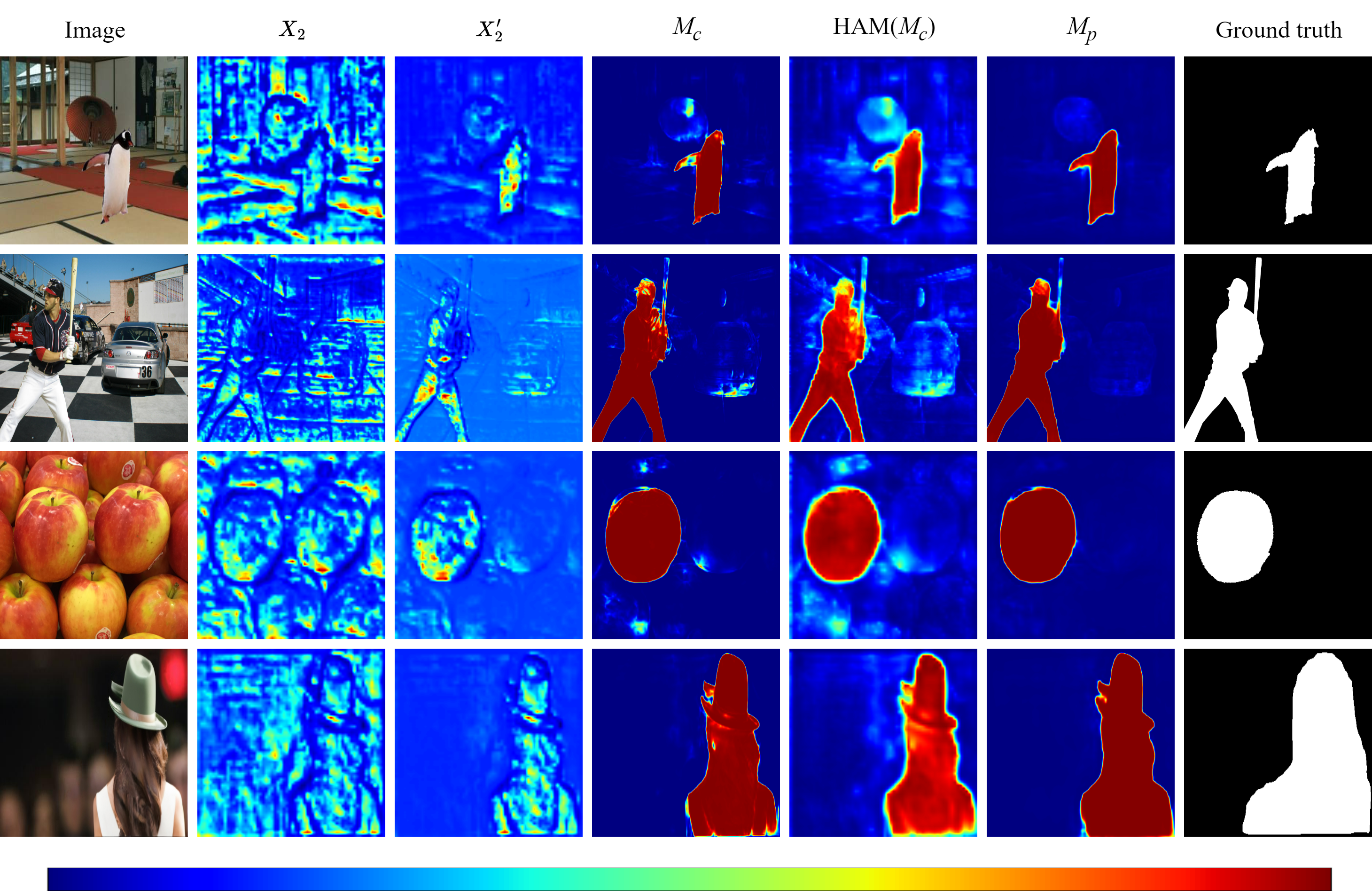}
% \captionsetup{skip=-2pt}
\vspace{-0.5cm}
\caption{Effectiveness of FEB verified by the visualized feature maps $X_2$, ${{X'}_2}$, ${M_c}$, HAM($M_c$) and ${M_p}$ on example images. The redder indicates higher responses. From top to bottom, the forgery images are from CASIAv1, NIST16, Coverage and AutoSplice, respectively.} \label{fig_2}
\end{figure}

\subsubsection{Feedback Enhancement Branch}
The feedback enhancement branch (FEB) is proposed to further improve the forensic representation ability. As plotted in Fig. \ref{fig_1} bottom, the FEB subnetwork contains two MiT blocks followed by a CSPM and an MLP decoder. To align with the standard four encoder stages of a typical transformer network, FEB is connected to the second MiT block of the coarse branch. Specifically, the coarse localization map ${M_c}$ generated by CLB is passed through HAM \cite{wu2019cascaded}, then multiplied with the CSPM-enhanced ${X_2}$ to yield the input of FEB ${X'_2}$ as
\begin{equation}
\label{eq4}
{X'_2} = {\rm{HAM}}\left( {\left( {{M_c}} \right)} \right) \otimes {\rm{CSPM}}\left( {{X_2}} \right)
\end{equation}
where  $\otimes$ denotes the element-wise multiplication.The HAM is defined as
\begin{equation}
\label{eq5}
% {\rm{HAM}}\left( {{M_c}} \right) = MAX\left[ \begin{array}{l}
% {f_n}\left( {{\rm{Con}}{{\rm{v}}_{\rm{g}}}\left( {{\rm{Con}}{{\rm{v}}_3}\left( {{M_c}} \right)} \right)} \right),{\rm{ }}\\
% {\rm{Con}}{{\rm{v}}_3}\left( {{M_c}} \right)
% \end{array} \right]
{\rm{HAM}}\left( {{M_c}} \right) = MAX\left[ {{f_n}\left( {{\rm{Con}}{{\rm{v}}_{\rm{g}}}\left( {{{\hat M}_c}} \right)} \right),{\rm{ }}{{\hat M}_c}} \right]
\end{equation}
where ${\hat M_c}$ is the downsampled ${M_c}$ via a convolutional layer with kernel size 3, stride 2, and padding 1. ${\rm{Con}}{{\rm{v}}_{\rm{g}}}$ is the convolution with Gaussian kernel and zero bias. ${f_n}\left(  \cdot  \right)$ is a normalization function to make the value of Gaussian-blurred map between 0 and 1. $MAX\left[ { \cdot ,  \cdot } \right]$ is a maximum function. The HAM is mainly aimed at enlarging the coverage area of the coarse map ${M_c}$ to reduce the loss of filtered information.
Two transformer blocks are stacked to ensure model efficiency, enhancing the learning and extraction of the features ${X'_3}$, ${X'_4}$ from ${X'_2}$. The forgery localization decoder ${D_2}$ of FEB, which shares the same structure with ${D_1}$, is configured to predict the final localization map ${M_p}$ as
\begin{equation}
\label{eq6}
{M_p} = \sigma \left( {{\rm{Upsample}}\left( {{D_2}\left( {{{X'}_2}, {{X'}_3}, {{X'}_4}} \right)} \right)} \right)
\end{equation}
where the multi-scale features ${{X'}_2}$, ${{X'}_3}$, ${{X'}_4}$ in the FEB are aggregated to infer the final decision map.

To verify the reasonability of FEB design, the input related feature maps ${X_2}$, ${{X'}_2}$, ${M_c}$, HAM$(M_c)$ and the output ${M_p}$ of FEB on example images are visualized in Fig. \ref{fig_2}. Owing to the feedback enhancement from CLB, apparent improvement can be seen from ${X_2}$ to ${{X'}_2}$, where the forged regions become more prominent with less false detection errors. Following the application of HAM, $M_c$ is blurred to extend the coverage of coarse localization maps, thereby ensuring the inclusion of more pertinent information. This approach helps prevent the significant impact of inaccurate coarse localization maps on the final results. Compared with the coarse localization map ${M_c}$, the fine one ${M_p}$ yielded by FEB enjoys more precise and consistent responses both inside and outside the forged regions.

\subsubsection{Contextual Spatial Pyramid Module}
\begin{figure*}[!t]
\centering
\includegraphics[width=\linewidth]{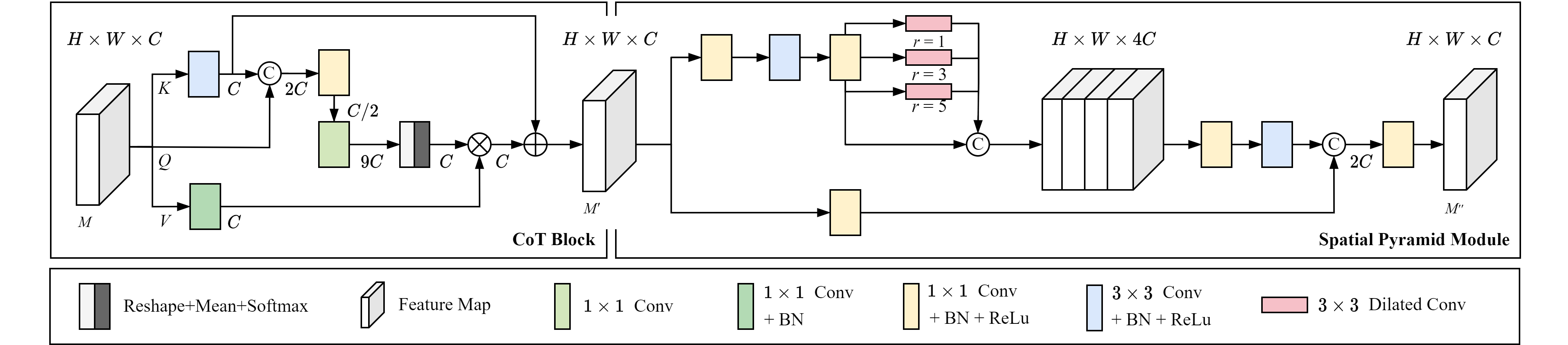}
\caption{Detailed structures of CSPM. The input feature $M$ passes the CoT block for exploring contextual information and is then enhanced by a spatial pyramid of dilated convolutions to output $M''$. $\oplus$, $\otimes$, $\copyright$ denote element-wise addition, multiplication, and concatenation, respectively.}
\label{fig_3}
\end{figure*}

To enhance the features of transformer blocks, we propose a new contextual spatial pyramid module (CSPM) which includes two main components as shown in Fig. \ref{fig_3}. Firstly, a Contextual Transformer (CoT) block \cite{li2022contextual} is employed to enhance the transformer backbone architecture. Such a block leverages the contextual information among input keys to guide self-attention learning. Secondly, a spatial pyramid of dilated convolutions is utilized to acquire the feature representations with distinct receptive fields. It is significant to improve the robustness of localizers against the variability of object regions and forgery image resolutions. Existing approaches often rely on spatial pooling pyramids \cite{wang2023yolov7} for this purpose. However, recent studies indicate that the pooling may be unsuitable for capturing the sensitivity to subtle signals, as it strengthens content and suppresses weak noises \cite{boroumand2018deep}. To attenuate such deficiency, here we take advantage of the dilated convolutions with different rates to enlarge the receptive field for covering larger contexts, thus capturing more subtle tampering traces at multiple scales.

The detailed structure of CSPM is illustrated in Fig. \ref{fig_3}. To be specific, assume the input feature $M\in\mathbb{R}^{C\times H\times M}$. First, a $3\times3$ group convolution is used to extract static contextual information between local neighbors. The contextual keys ($K$) and queries ($Q$) are concatenated and the attention matrix is calculated through two consecutive $1\times1$ convolutions. The dynamic contextual representation is obtained by aggregating $V$ after $1\times1$ convolution. The output ${M'}$ is the fusion of such static and dynamic context features. That is,
\begin{equation}
\label{eq7}
\begin{split}
M' = &{\rm{Con}}{{\rm{v}}_{3 \times 3}}\left( M \right) \oplus \\
&{\rm{Concat}}\left[ {{\rm{Con}}{{\rm{v}}_{3 \times 3}}\left( M \right),Q} \right]{W_\theta }{W_\delta } \otimes \left( {{W_\phi }V} \right)
\end{split}
\end{equation}
where ${W_\theta }$ , ${W_\delta }$ , ${W_\phi }$  are the involved $1\times1$ convolutions. Subsequently, the feature enhancement process includes two branches. One branch enforces regular $1\times1$ convolution, and the other applies multiple parallel filters with different rates for capturing multi-scale features. The results of such two pipelines are merged and reshaped to $H\times$$W\times$$C$ by a  $1\times1$ convolution, resulting in the final enhanced feature $M''$ . 

\subsection{Loss Function}
The proposed ProFact network is trained end-to-end with a total loss function ${L_{\rm{total}}}$ including two branch terms, which are the losses between the predicted coarse (${M_c}$) or refined (${M_p}$) localization map and the ground truth $G$. That is,
\begin{equation}
\label{eq8}
{L_{\rm{total}}} = L\left( {{M_c},G} \right) + L\left( {{M_p},G} \right).
\end{equation}
To alleviate the quantity imbalance between the pristine and forged pixels in training samples, the combined loss consisting of focal loss \cite{lin2017focal} and dice loss \cite{milletari2016v} is adopted as
\begin{equation}
\label{eq9}
L\left( {M_{\cdot } ,G} \right) = \lambda {L_{\rm{focal}}}\left( {M_{\cdot } ,G} \right) + \left( {1 - \lambda } \right){L_{\rm{dice}}}\left( {M_{\cdot } ,G} \right){\rm{ }}
\end{equation}
where $\lambda $ is empirically set to 0.5. The focal loss is defined as
\begin{equation}
\label{eq10}
\begin{split}
{L_{\rm{focal}}}\left( {{y}, {\hat{y}}} \right) =  & - \sum {\alpha {{\left( {1 - {\hat{y}}} \right)}^\gamma }} {y}\log \left( {{\hat{y}}} \right)\\
                         & - {\sum {\left( {1 - \alpha } \right){\hat{y}}} ^\gamma }\left( {1 - {y}} \right)\log \left( {1 - {\hat{y}}} \right)
\end{split}
\end{equation}
where  $y$ and $\hat{y}$ are the ground truth and prediction label for each pixel of training sample images, respectively. Following the prior work \cite{lin2017focal}, the weights $\alpha$ and $\gamma$ are set as 0.5 and 2, respectively. The dice loss is defined as
\begin{equation}
\label{eq11}
{L_{\rm{dice}}}\left( {{y},{\hat{y}}} \right) = 1 - \frac{{2 \cdot \sum {\left( {{y} \cdot {\hat{y}}} \right)} }}{{\sum {\left( {{y} + {\hat{y}}} \right)} }}
\end{equation}

\subsection{Generation of Realistic Forged Training Images}
The deep learning-based forgery localizers typically require large-scale synthesized training samples. Existing methods (refer to Part C of Section II) have not fully addressed the realism of simulating forged images. As illustrated in Fig. 4, we propose a simple yet effective automatic composition method named MBH to create large-scale realistic forged training images. Primarily, a COCO \cite{lin2014microsoft} image is randomly selected as the foreground, and its included objects are chosen as target regions. A background image is also randomly selected for splicing, whereas the foreground image itself is used for copy-move. 

\begin{figure}[!tb]
\centering
\includegraphics[width=\linewidth]{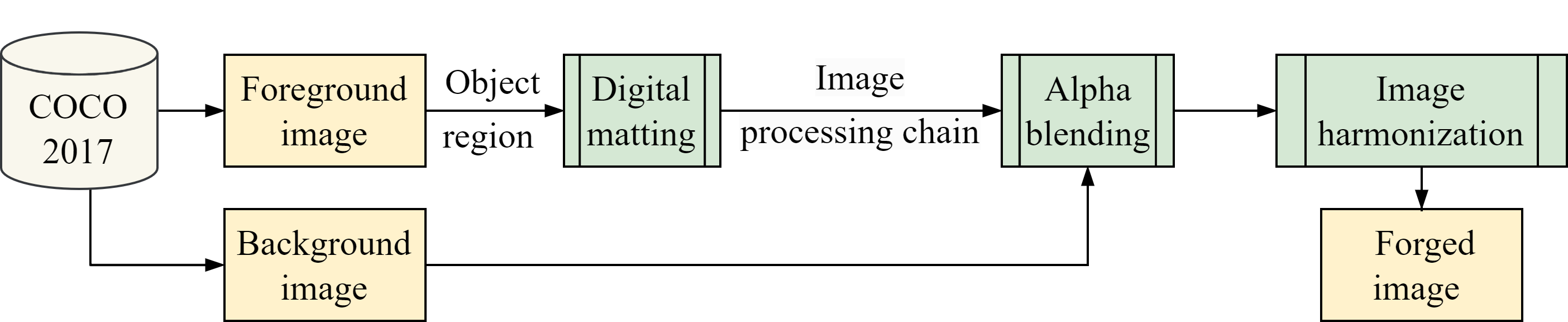}
\caption{Proposed realistic training samples generation method. It successively includes digital matting, the processing chain with scaling, rotation, flipping and deformation, alpha blending, and harmonization.}
\label{fig_4}
\end{figure}

{\bf{Matting.}} A trimap corresponding to the target region is obtained by applying dilation and erosion to the annotations in the COCO dataset. With such a trimap, the state-of-the-art digital matting algorithm, i.e., SIM \cite{sun2021semantic} is called to extract refined alpha mattes, which enjoy fine object boundaries and appropriate transparency.  

{\bf{Blending.}} A processing chain is applied to the segmented object regions for simulating the common image transformation, which is a random combination of scaling, rotation, flipping and deformation. The parameter settings for such manipulations are listed in Table \ref{tab:table2}. To reduce unnatural splicing boundaries, alpha blending \cite{porter1984compositing} is adopted to seamlessly blend the object into the background. Note that the target region is required to be within [0.5\%, 50\%] of the global image size, and spliced at a random position.  

\begin{table}[!tb]
\caption{Parameters of manipulations.\label{tab:table2}}
\centering
\begin{tabular}{ll}
\toprule
       Manipulation Type & Manipulation Parameters \\ \midrule
        Scaling & range = [0.5, 2.0], bicubic interpolation kernel \\ \midrule
        Rotation & angle = [-180, 180] \\ \midrule
        \multirow{2}{*}{Flipping} & vertical/horizontal/ \\
        & both horizontal and vertical flipping \\ \midrule
        Deformation & range = [0.5, 2.0], bicubic interpolation kernel \\  \bottomrule
\end{tabular}
\end{table}

{\bf{Harmonization.}} Finally, a classical image harmonization algorithm \cite{lalonde2007using} is applied randomly to enhance the consistency of color perception between the target and background regions. The resulting realistic forgery images comprise the synthesized dataset denoted by $D_{\rm{MBH-COCO}}$.  

\begin{figure*}[!t]
\centering
\includegraphics[width=0.95\linewidth]{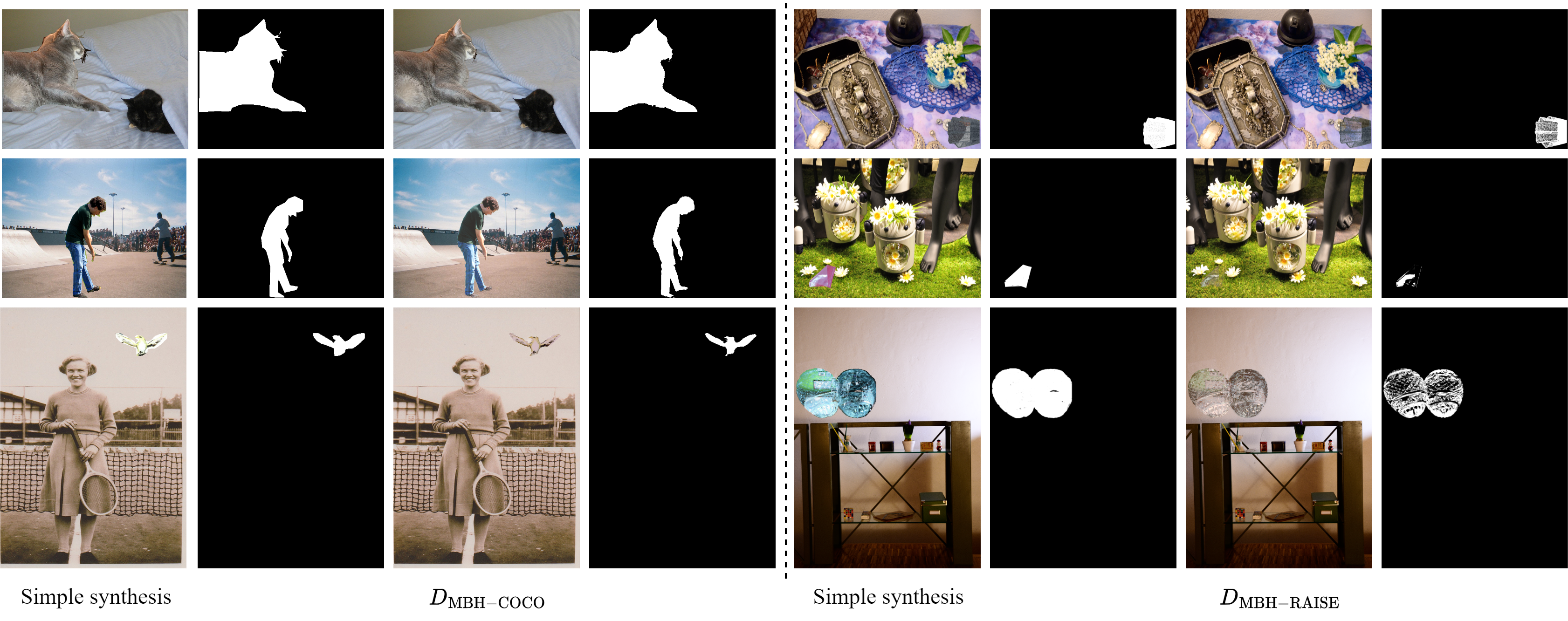}
\caption{Example forged images generated by simple synthesis and MBH methods, and their corresponding ground-truths. Zoom in for details.}
\label{fig_5}
\end{figure*}

Furthermore, several specialized matting datasets are used to collect more precise target regions without annotation refinement. Since such foreground images have higher spatial resolutions than the COCO ones, the background images with matched sizes would be randomly selected from the RAISE dataset \cite{dang2015raise}. Then another realistic forgery image set, called $D_{\rm{MBH-RAISE}}$, is created by applying the same processing to such target regions and background images. $D_{\rm{MBH-RAISE}}$ enjoys more comprehensive boundary and texture refinements than $D_{\rm{MBH-COCO}}$.

Fig. \ref{fig_5} illustrates some forged images generated by simple synthesis and MBH methods, along with their corresponding pixel-level ground-truths. The simple synthesis refers to the direct splicing between target regions and background images. Although the MBH pays less attention to object placement and semantic coherence to limit the computational cost, the created images exhibit higher realism, such as natural splicing boundaries and color consistency. As a result, our proposed method can well simulate the tampering operations, and generate sufficient and realistic forged training samples.

\subsection{Two-Stage Training Protocol}
As progressive learning, the training of our ProFact network covers two stages. The network is first trained on the dataset $D_{\rm{MBH-COCO}}$, and then fine-tuned on $D_{\rm{MBH-RAISE}}$. Such fine-tuning allows the network to fully leverage the network weights learned in the first stage and mitigate the risk of overfitting. Through refining the features learned in the previous stage, the network can progressively learn complex features on forensic clues. Specially, in the second stage, the sizes of the image patches fed into ProFact are increased to adapt to the larger spatial resolution of $D_{\rm{MBH-RAISE}}$ images for reducing information loss. Such adaption can improve the robustness against the dimension variation of test images in real-world scenarios, and facilitate learning more extensive context information and global structures.

\section{EXPERIMENTS}
In this section, we first introduce training and testing datasets, evaluation metrics, and training details. We present both quantitative and qualitative evaluation results on nine public datasets in comparison with seven state-of-the-art methods, ablation studies and robustness assessments.

\subsection{Experimental Setup}
\subsubsection{Datasets}

\textbf{Experiment 1.} We train the model using the CASIAv2 dataset and compare its performance with other state-of-the-art methods trained on the same dataset for a fair evaluation. Then we test our model on ten datasets aligned with the latest method\cite{kong2023pixel}.

\textbf{Experiment 2.} Table \ref{tab:table3} summarizes the datasets used for training and testing in this main experiment. The generated datasets $D_{\rm{MBH-COCO}}$ and $D_{\rm{MBH-RAISE}}$ including forged images and their corresponding ground-truths are used for training the ProFact network. In the first stage of training, the $D_{\rm{MBH-COCO}}$ with 30,000 splicing, 15,000 copy-move and 5,000 inpainting images is involved. Both the splicing and copy-move samples are generated using our proposed MBH method, while the inpainting ones are randomly selected from the public dataset in \cite{liu2022pscc}. In the second stage, the network is fine-tuned on the $D_{\rm{MBH-RAISE}}$ with 5,000 generated images, which splice 2,000 foreground images from the public matting datasets AIM \cite{li2021deep}, Distinctions \cite{qiao2020attention}, PPM \cite{ke2022modnet}, RWP \cite{yu2021mask}, and SIMD dataset \cite{sun2021semantic} to 5,000 background images from RAISE dataset \cite{dang2015raise}. In each stage, a validation set with 1/10 the size of the training set is also built. 
To comprehensively evaluate various forgery types, extensive tests are conducted on ten standard datasets. Such datasets are widely used in literatures and contain common tampering types, such as splicing, copy-move and inpainting. Specifically, the Korus contains 220 high-quality realistic forged images, which cover object insertion and removal as well as subtle content changes, such as reshaped shadows and reflections. The IMD collects 2,010 manipulated images from the Internet. The AutoSplice \cite{jia2023autosplice} includes 3,621 locally synthetic images generated by the DALL-E2 language-image model.
\subsubsection{Metrics}
Similar to previous works \cite{wu2022robust}, the accuracy of pixel-level forgery region localization is measured by F1 and IoU defined as
\begin{equation}
\label{eq12}
{\rm{F1}} = \frac{{{\rm{2}} \times {\rm{TP}}}}{{{\rm{2}} \times {\rm{TP}} + {\rm{FP}} + {\rm{FN}}}}
\end{equation}
\begin{equation}
\label{eq13}
{\rm{IoU}} = \frac{{{\rm{TP}}}}{{{\rm{TP}} + {\rm{FP}} + {\rm{FN}}}}
\end{equation}
where TP and FN denote the numbers of correctly and incorrectly classified forged pixels in a test forgery image, respectively. TN and FP denote those corresponding to pristine pixels. The fixed threshold 0.5 is adopted to binarize the predicted tampering probability map for computing F1 and IoU scores. The localization accuracy on a test dataset refers to the average metric value of all images in the dataset.

\begin{table}[!tb]
\caption{Summary of datasets used for training and testing in Experiment {\scriptsize{2}}. SP, CM, IN, AI are short for splicing, copy-move, inpainting and local AI editing, respectively.} \label{tab:table3}
\centering
\resizebox{\columnwidth}{!}{
\begin{tabular}{ll|cccc|l|l}
\toprule
\multirow{2}{*}{Dataset} & \multirow{2}{*}{Images} & \multicolumn{4}{c|}{Forgery Types} & \multirow{2}{*}{Resolution} & \multirow{2}{*}{Format} \\  \cmidrule{3-6}
&    & SP   & CM   & IN   & AI   & (Average)    &    \\ \midrule
\multicolumn{8}{l}{\emph{Training}}    \\
$D_{\rm{MBH-COCO}}$ & 50,000 & $\checkmark$ & $\checkmark$ & $\checkmark$ &  & $640 \times 576$     &  PNG, TIF    \\
$D_{\rm{MBH-RAISE}}$    & 5,000  & $\checkmark$ & $\checkmark$ & &  & $4608 \times 3712$    & PNG    \\ \midrule
\multicolumn{8}{l}{\emph{Testing}}    \\
Columbia\cite{hsu2006detecting}  & 160   & $\checkmark$ & & &   & $1152 \times 768$ & PNG    \\
CASIAv1\cite{dong2013casia}   & 921   & $\checkmark$ & $\checkmark$ & &     & $384 \times 256$   & JPEG   \\
NIST16\cite{guan2019mfc}   & 564   & $\checkmark$ & $\checkmark$ & $\checkmark$ &   & $5616 \times 3744$    & JPEG   \\
DSO\cite{de2013exposing}   & 100    & $\checkmark$ & & &   & $2048 \times 1536$        & PNG    \\
IMD\cite{novozamsky2020imd2020}    & 2,010   & $\checkmark$ & $\checkmark$ & $\checkmark$ &   & $1920 \times 1200$ & JPEG   \\
Korus\cite{korus2016evaluation}  & 220    & $\checkmark$ & $\checkmark$ & $\checkmark$ &   & $1920 \times 1080$   & TIF     \\
Coverage\cite{wen2016coverage}   & 100    & & $\checkmark$ & &    & $480 \times 400$    & TIF    \\
Wild\cite{huh2018fighting}    & 201  & $\checkmark$ & $\checkmark$ & $\checkmark$ &   & $1600 \times 900$     & JPEG   \\
CocoGlide\cite{guillaro2023trufor}   & 512   & & & & $\checkmark$     & $256 \times 256$      & PNG  \\ 
AutoSplice\cite{jia2023autosplice}  & 3,621   & & & & $\checkmark$   & $384 \times 304$     & JPEG  \\ \bottomrule
\end{tabular}}
\end{table}

\subsubsection{Implementation details}
The proposed ProFact network is implemented by PyTorch. All experiments are conducted on an A800 GPU server. We initialize the backbone network MiT-B3 \cite{xie2021segformer} with pre-trained ImageNet weights and adopt AdamW as the optimizer. The learning rate starts from 1e-4 and decreases by the cosine annealing strategy. We set the batch size as 16, and the training period as 50 epochs. In the second stage, such parameters are reset as 4, 1e-5, and 5, respectively. We choose the best model in terms of the IoU score on the validation set. 
The following augmentations are applied to input RGB training images: resizing within the range of [0.5, 2.0], cropping a fixed-size block ($512\times512$, $1024\times1024$ for the two stages respectively) with the area proportion from 5\% to 75\%, horizontal flipping with a probability of 0.5, and JPEG compression with a quality factor ranging from 70 to 95.

\begin{table*}[!ht]
\caption{Localization performance F1/IoU[\%] comparison of different methods trained on CASIAv2 dataset and tested on other different datasets. The best results are in \textbf{bold} and the second best results are \underline{underlined}.}\label{tab:table4}
\centering
\resizebox{\linewidth}{!}{
\begin{tabular}{llccccccccccc}
\toprule
Method & Venue  & Columbia & CASIAv1 & NIST16 & DSO & IMD & Korus  & Coverage  & Wild &DEF-12k & IFC & Average \\ \midrule
MFCN\cite{salloum2018image}     & JVCIP18 & 18.4/12.3 & 34.6/29.1  & 24.3/19.3  & 15.0/10.3 & 17.0/12.4 & 11.8/8.3 & 14.8/10.0 & 16.1/11.2  & 6.7/5.0  & 9.8/7.4  & 16.9/12.5    \\
Mantra\cite{wu2019mantra}  & CVPR19  & 45.2/30.1 & 18.7/11.1   & 15.8/9.8   & 25.5/15.3  & 16.4/9.8   & 11.0/6.1 & 23.6/13.9 & 31.4/20.1 & 6.7/3.9    & 11.7/6.8  & 20.6/12.7    \\
H-LSTM\cite{bappy2019hybrid} & TIP19   & 14.9/9.0 & 15.6/10.1   & \textbf{35.7}/\textbf{27.6}  & 14.2/8.4  & 20.2/13.1   & 14.3/9.4 & 16.3/10.8 & 17.3/10.6 & 5.9/3.7  & 7.4/4.7  & 16.2/10.7    \\
SPAN\cite{hu2020span}  & ECCV20 & 50.3/39.0 & 14.3/11.2    & 21.1/15.6  & 8.2/4.9  & 14.5/10.0  & 8.6/5.5 & 14.4/10.5  & 19.6/13.2  & 3.6/2.4   & 5.6/3.7   & 16.0/11.6    \\
ViT-B \cite{dosovitskiy2020image}     & ICLR21  & 21.7/16.4 & 28.2/23.2   & 25.4/19.7    & 16.9/12.1  & 15.4/19.2    & 17.6/13.0 & 14.2/10.1 & 20.8/15.2  & 6.2/4.5    & 7.1/5.1  & 17.4/13.9    \\
Swin-ViT \cite{liu2021swin}  & ICCV21  & 36.5/29.7 & 39.0/35.6   & 22.0/16.7  & 18.3/13.2 & 30.0/24.3  & 13.4/10.3 & 16.8/12.4 & 26.5/21.4 & 15.7/12.9  & 10.2/7.8  & 22.8/18.4    \\
PSCC \cite{liu2022pscc}    & TCSVT22 & 50.3/36.0 & 33.5/23.2  & 17.3/10.8   & \underline{29.5}/18.5  & 19.7/12.0   & 11.4/6.6 & 22.0/13.0 & 30.3/19.3 & 7.2/4.2    & 11.4/6.7  & 23.3/15.0   \\
MVSS++ \cite{dong2022mvss} & TPAMI22 & 66.0/57.3 & 51.3/39.7    & \underline{30.4}/\underline{23.9}     & 27.1/18.8  & 27.0/20.0   & 10.2/7.5 & \textbf{48.2}/\textbf{38.4} & 29.5/21.9  & 9.5/7.6  & 8.0/5.5  & 30.7/24.1    \\
CAT \cite{kwon2022learning}   & IJCV22  & 20.6/14.0 & 23.7/16.5   & 10.2/6.2   & 17.5/11.0  & 25.7/18.3   & 8.5/4.9 & 21.0/14.1 & 21.7/14.4  & \textbf{20.6}/\textbf{15.2}    & 9.9/6.2   & 17.9/12.1    \\
EVP \cite{liu2023explicit}   & CVPR23  & 27.7/21.3 & 48.3/42.1   & 21.0/16.0   & 6.0/4.3  & 23.3/18.3  & 11.3/8.4 & 11.4/8.3  & 23.1/18.2  & 9.0/7.0   & 8.1/6.2  & 18.9/15.0    \\
PIM \cite{kong2023pixel}     & arxiv23 & \underline{68.0}/\underline{60.4} & \underline{56.6}/\underline{51.2}     & 28.0/22.5     & 25.3/\underline{19.4}  & \underline{41.9}/\underline{34.0}   & \underline{23.4}/\underline{18.2} & 25.1/18.8 & \textbf{41.8}/\textbf{33.8} & \underline{16.7}/\underline{13.3}        & \underline{15.5}/\underline{11.9}  & \underline{34.2}/\underline{28.4}     \\
ProFact (ours)    & -     & \textbf{71.6}/\textbf{65.0} & \textbf{72.3}/\textbf{65.8}    & 23.6/15.9    & \textbf{40.6}/\textbf{32.6}  & \textbf{45.4}/\textbf{38.2}    & \textbf{31.4}/\textbf{24.5} & \underline{26.7}/\underline{22.4} & \underline{41.2}/\textbf{33.8} & 15.1/12.4    & \textbf{15.8}/\textbf{12.5}  & \textbf{38.4}/\textbf{32.3}   \\ \bottomrule
\end{tabular}}
\end{table*}

\begin{table*}[!tb]
\caption{Image forgery localization performance F1/IoU[\%] comparison of different methods on ten test datasets. "-", "${\dagger}$" mean inclusion of the dataset and CAISAv2 in training, respectively. The best results are in \textbf{bold} and the second best results are \underline{underlined}.} \label{tab:table5}
\centering
\resizebox{\linewidth}{!}{
\begin{tabular}{llccccccccccccc}
\toprule
Method & Venue  & Columbia & CASIAv1 & NIST16 & DSO  & IMD  & Korus & Coverage & Wild & CocoGlide & AutoSplice   & Average \\ \midrule
Noiseprint \cite{cozzolino2019noiseprint} & TIFS19 & 36.4/26.2 & 13.0/7.4 & 12.2/8.1 & 33.9/25.3 & 17.9/12.0 & 14.7/9.0  & 14.7/8.7  & 16.7/11.0 & 23.8/15.1  & 33.0/21.5 & 21.6/14.4    \\
ManTra \cite{wu2019mantra}  & CVPR19    & 35.6/25.8 & 13.0/8.6    & 9.2/5.4  & 33.2/24.3 & 18.3/12.4 & 17.9/11.8  & 27.5/18.6  & 15.6/11.0 & 33.4/25.1      & 18.2/12.2 & 22.2/15.5    \\
DFCN \cite{zhuang2021image}   & TIFS21    & 38.1/24.9 & 8.3/6.1     & -/-    & \underline{68.4}/\underline{57.9} & 17.3/11.4 & 10.8/6.1  & 18.5/10.6  & 23.5/16.2 & 34.5/25.2      & 64.6/52.3 & 31.6/23.4    \\
MVSS \cite{dong2022mvss}  & PAMI22   & 68.4/59.6 & 45.1/39.7$^{\dagger}$    & 29.4/24.0 & 27.1/18.8 & 26.0/20.0 & 9.5/6.7   & 44.5/37.9  & 29.5/21.9 & 35.6/27.5      & 33.3/24.1 & 34.8/28.0    \\
PSCC \cite{liu2022pscc}  & TCSVT22     & 61.5/48.3 & 46.3/41.0  & 18.7/13.5 & 41.1/31.6 & 15.8/12.1 & 10.2/5.8  & 44.4/33.6  & 10.8/8.1 & \underline{42.2}/\underline{33.3}      & 60.4/49.0 & 35.1/27.6    \\
OSN \cite{wu2022robust}   & TIFS22     & 71.3/61.4 & 50.9/46.5    & 33.1/25.2 & 44.5/31.7 & \underline{49.1}/\underline{39.2} & 29.9/21.8  & 26.0/17.6  & 50.5/39.4 & 26.5/20.7   & 50.9/39.5 & 43.3/34.3    \\
CAT \cite{kwon2022learning}  & IJCV22     & 79.3/74.6 & \textbf{71.0}/\textbf{63.7}$^{\dagger}$     & 30.2/23.5 & 47.9/40.9 & -/- & 6.1/4.2  & 28.9/23.0  & 34.1/28.9 & 36.3/28.8  & \textbf{86.2}/\textbf{78.0} & 46.7/40.6    \\
HiFi \cite{guo2023hierarchical}  & CVPR23    & \textbf{86.9}/78.9 & 10.8/8.8    & 29.6/21.9 & 29.9/21.1 & 10.6/7.5 & 8.2/4.9   & 39.8/29.8  & 10.0/6.8 & 22.9/16.6      & 17.8/11.5 & 26.6/20.8    \\
TBFormer \cite{liu2023tbformer} &SPL23 & 83.6/79.8  &\underline{69.6}/\underline{63.6}$^{\dagger}$  &34.1/28.6  &5.9/4.6  &40.1/33.3  &19.7/14.9  &\underline{52.2}/\textbf{45.5}  &32.0/26.6  &23.9/19.5  &14.1/10.1  &37.5/32.7  \\
TruFor \cite{guillaro2023trufor} &CVPR23 &79.8/74.0 &\underline{69.6}/\underline{63.2}$^{\dagger}$  & \textbf{47.2}/\textbf{39.6} &\textbf{91.0}/\textbf{86.5} &-/-  &\textbf{37.7}/\textbf{29.9} &\textbf{52.3}/\underline{45.0} &\underline{61.2}/\underline{51.9} &35.9/29.1 &\underline{65.7}/\underline{54.7} &\textbf{60.0}/\textbf{52.7} \\
ProFact (ours)      & -    & \underline{84.5}/\textbf{82.7} & 56.4/50.2    & \underline{43.1}/\underline{36.2} & 46.4/36.1 & \textbf{53.8}/\textbf{45.9} & \underline{31.5}/\underline{25.0}  & 51.1/43.1  & \textbf{64.5}/\textbf{56.1} & \textbf{69.2}/\textbf{61.3}   & 65.5/54.4 & \underline{56.6}/\underline{49.1}  \\ \bottomrule
\end{tabular}}
\end{table*}

\begin{table*}[!tb]
\caption{Comprehensive comparison with the performance of CAT and TruFor over ten test datasets. "-", "${\dagger}$" mean inclusion of the dataset and CASIAv2 in training, respectively.} \label{tab:table6}
\centering

\begin{tabular}{ll|cccccccccc|cc|cc}
\toprule
\multirow{2}{*}{Method} & \multirow{2}{*}{Training} & \multicolumn{10}{c|}{Testing Datasets (IoU[\%])} & \multirow{2}{*}{Avg} & \multirow{2}{*}{Std} & \multirow{2}{*}{FLOPs} & \multirow{2}{*}{Inference} \\ \cmidrule{3-12}
&Images & Col &CASv1 &NIST & DSO & IMD & Korus & Cov & Wild & Glide & Auto &(\%) &(\%) &(G) &Time (ms) \\ \midrule
CAT & 826K     & 74.6 & 63.7$^{\dagger}$  & 23.5 & 40.9 &  & 4.2  & 23.0  & 28.9 & 28.8  & 78.0 & 40.6 &24.2 &134.1 & 45.3   \\
TruFor &826K &74.0 & 63.2$^{\dagger}$  & \textbf{39.6} & \textbf{86.5} & - & 29.9 & 45.0 & 51.9 & 29.1 & 54.7 & 52.7 &18.3 &221.8&73.9 \\
ProFact   & 55K   & \textbf{82.7} & 50.2   & 36.2 & 36.1 & \textbf{45.9} & 25.0  & 43.1  &56.1 & \textbf{61.3}  & 54.4 & 49.1  &15.2 &118.5 &37.2  \\ 
ProFact$^{\star}$   & 81K   & 65.7 & \textbf{73.5}$^{\dagger}$  & 36.7 & 41.3 & - & \textbf{32.7}  &\textbf{44.1}  & \textbf{67.6} &57.9   & \textbf{57.4} & \textbf{53.0} &\textbf{13.9} &118.5 &37.2   \\ \bottomrule
\end{tabular}
\end{table*}

\begin{figure*} [!tb]
\centering
\includegraphics[width=\linewidth]{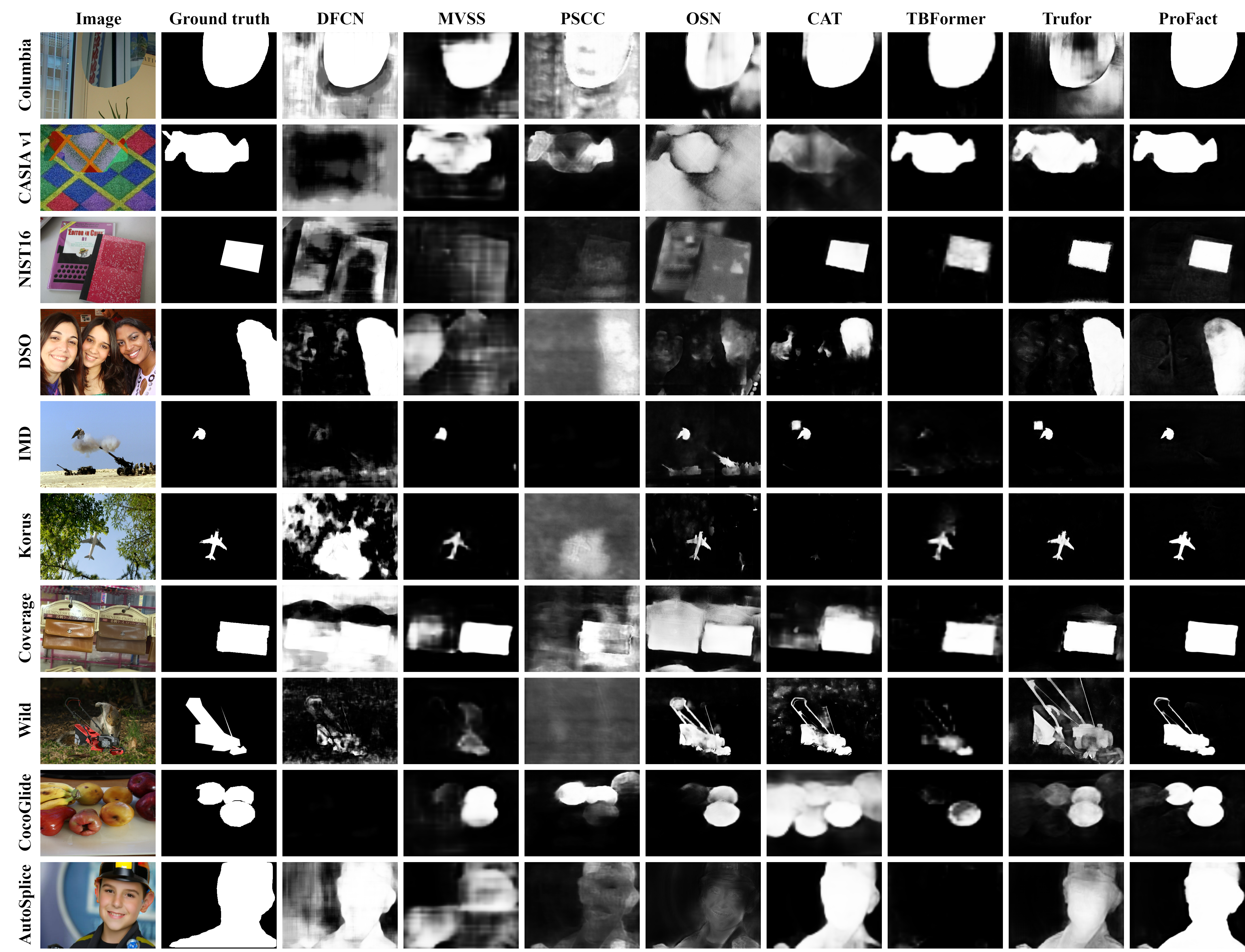}
% \captionsetup{skip=0pt}
\vspace{-0.5cm}
\caption{Localization results of different methods on example forged images from various datasets. From left to right: forged image, ground truth, and the localization maps from seven prior and our ProFact methods.}\label{fig_6}
\end{figure*}

\subsection{Comparison with Other Localization Algorithms}

\subsubsection{Experiment 1}
To validate the effectiveness of our model, we train our model on CASIAv2 dataset and compare its performance with the other state-of-the-art methods for a fair evaluation. It is noted that the performance metrics reported in Table \ref{tab:table4} are based on the results from the latest work \cite{liu2023tbformer}, where all the compared forgery localization models have been retrained on the CASIAv2 dataset.

As indicated in the results, our method achieves the highest localization accuracy across ten datasets. Comparing with the MVSS++\cite{dong2022mvss}, our method achieves 7.7\% and 8.2\% increases for the average F1 and IoU scores, respectively. Our method also outperforms the latest model PIM\cite{kong2023pixel} with 4.2\% and 3.9\% increases on such two measurements. Such results validate the effectiveness and advantage of our proposed ProFact method. 

\subsubsection{Experiment 2}
To ensure a fair comparison, we only consider the image forgery localizers with publicly available code or pre-trained models. Specifically, we compare the ProFact with seven state-of-the-art localization methods, including Noiseprint \cite{cozzolino2019noiseprint}, ManTra \cite{wu2019mantra}, DFCN \cite{zhuang2021image}, MVSS \cite{dong2022mvss}, OSN \cite{wu2022robust}, PSCC \cite{liu2022pscc}, CAT \cite{kwon2022learning}, HiFi \cite{guo2023hierarchical}, TBFormer \cite{liu2023tbformer} and TruFor \cite{guillaro2023trufor}. Table \ref{tab:table5} shows the evaluation results for the pixel-level localization performance of different methods. The ranking for each method is also indicated in parentheses, with the three best ones highlighted in red. Note that ManTra and PSCC fail to deal with some rather high-resolution images in NIST16, IMD, and AutoSplice due to GPU memory limitations. Following the previous work \cite{kwon2022learning}, such images are cropped to $2048 \times 1440$ pixels for testing.

Overall, our method achieves the best IoU scores across four datasets, and outperforms the third-ranked method CAT (published on IJCV, 2022) by 9.9\% and 8.5\% in average F1 and IoU scores, respectively. It is worth highlighting that our method gains outstanding performance on the real-world dataset ‘Wild’ and the highly realistic tampering dataset ‘Korus’. Furthermore, our method performs well on the locally AI-generated datasets, i.e., CocoGlide and AutoSplice, which is unseen in the training phase. Such a result further presents the powerful cross-operation generalization ability of ProFact. These advantages should be rather significant for pushing forward the forensic localizers into real-world applications.

While our method exhibits slightly lower performance in terms of average F1 and IoU scores compared to TruFor, it still achieves competitive performance. Note that CAT and TruFor utilize over 826K image samples for model training. Additionally, the training samples includes several standard datasets similar to the test set, such as CASIAv2, which may lead to biased optimistic test results. Given these factors, we discuss a detailed comparison with such two methods in Table \ref{tab:table6}. In alignment with TruFor and CAT, we use publicly available datasets, including CASIAv2\cite{dong2013casia}, IMD\cite{novozamsky2020imd2020}, and FantasticReality\cite{kniaz2019point}, to extend the automatically synthesized dataset for training. As the results shown in the last row of Table \ref{tab:table6}, ProFact$^{\star}$ shows improved performance across most datasets. The average IoU value across ten datasets is 12.4\% higher than CAT, and it is comparable to TruFor. Despite potential biases arising from the differences in training dataset sizes, our method still delivers superior performance with  higher computational efficiency, characterized by lower FLOPs and shorter inference times. Furthermore, ProFact exhibits a lower standard deviation (std) compared to TruFor and CAT, indicating more consistent and stable performance across different forgery datasets.

We also qualitatively compare the performance of different localization methods on example image forgeries. One sample is selected from each test dataset for encompassing the three popular tampering types, i.e., splicing, copy-move and inpainting, and local artificial intelligence (AI) editing. Fig. \ref{fig_6} shows the predicted pixel-level forgery localization maps of different methods on such example images. It can be seen that our method produces more accurate localization results for different types of forged images. In most cases, the other localizers can merely detect part of forged regions with more or less false alarms. The performance advantage of ProFact should be attributed to its ability to learn the powerful and discriminative forensic features in the manners of feedback and progressive, which benefits accurately identifying the forged regions in challenging scenarios.

\subsection{Ablation Studies}
In this section, we conduct extensive ablation studies to validate the effectiveness of our proposed ProFact network. Its different variants are retrained and tested to assess the impact of each component. The localization performance is compared on Columbia, CASIAv1, DSO, and AutoSplice. A summary of the involved sub-experiments is shown in Table \ref{tab:table7}.

\begin{table*}[!t]
\caption{Localization performance comparisons for ablation study. Metric values are in percentage. \label{tab:table7}}
\centering
\footnotesize
\begin{tabular}{lcccccccc}
\toprule
        \multirow{2}*{Feedback Enhancement} & CLB & $\checkmark$ & ~ & ~ & ~ & ~ & ~ & ~ \\ 
        & CLB+FEB & ~ & $\checkmark$ & $\checkmark$ & $\checkmark$ & $\checkmark$ & $\checkmark$ & $\checkmark$ \\ \midrule
        \multirow{4}*{Feature Enhancement} & w/o CSPM  & ~ & $\checkmark$ & ~ & ~ & ~ & ~ & ~\\ 
        & only CoT & ~ & ~ & $\checkmark$ & ~ & ~ & ~ & ~\\
        & CSPM (w/o dilated conv) & ~ & ~  & ~ & $\checkmark$ & ~ & ~ & ~\\ 
        & CSPM (with dilated conv) & $\checkmark$  & ~ & ~ & ~ & $\checkmark$ & $\checkmark$ & $\checkmark$ \\     \midrule
        \multirow{3}*{Training Dataset and Stages} & $D_{\rm{Simple-COCO}}$ (Single stage)  & ~ & ~ & ~ & ~ & $\checkmark$ & ~ & ~ \\ 
        & $D_{\rm{MBH-COCO}}$ (Single stage) & ~ & ~ & ~ & ~ & ~ & $\checkmark$ & ~ \\ 
        & $D_{\rm{MBH-COCO}}$ + $D_{\rm{MBH-RAISE}}$ (Two stages) & $\checkmark$ & $\checkmark$ & $\checkmark$ & $\checkmark$ & ~ & ~ & $\checkmark$ \\  \midrule
        \multirow{2}*{Columbia} & F1 & 73.6 & 69.8 &71.9 & 84.2 & \bf{92.2} & 74.3 & 84.5 \\ 
        & IoU & 70.6 & 66.0 &68.6 & 81.7 & \bf{90.3} & 72.3 & 82.7 \\ \midrule
        \multirow{2}*{CASIAv1} & F1 & 54.7 & 51.7 &56.4 & 52.7 & 29.7 & \bf{59.8} & 56.4 \\ 
        & IoU & 48.9 & 46.6 &50.5 & 46.1 & 26.3 & \bf{54.5} & 50.2 \\ \midrule
        \multirow{2}*{DSO} & F1 & 37.9 & 34.9  &30.4 & 36.3 & 16.0 & 41.1 & \bf{46.4} \\ 
        & IoU & 32.1 & 28.7 & 26.2 & 29.1 & 12.7 & 34.7 & \bf{36.1} \\ \midrule
        \multirow{2}*{AutoSplice} & F1 & 58.7 & 56.2 &57.9 & 54.7 & 10.5 & 51.2 & \bf{65.5} \\ 
        & IoU & 47.5 & 45.6 &47.7 & 44.0 & 8.1 & 41.8 & \bf{54.4} \\ \midrule
        \multirow{2}*{Average} & F1 & 56.2 & 53.2 &54.1 & 57.0 & 37.1 & 56.6 & \bf{63.2} \\ 
        & IoU & 49.8 & 46.7 &48.3 & 50.2 & 34.4 & 50.8  & \bf{55.9} \\ \bottomrule
\end{tabular}
\end{table*}

{\textbf{Influence of feedback enhancement.}} We train and test the proposed forgery localization network without the feedback enhancement branch (FEB), namely the CLB network. Compared with the intact ProFact network, the average F1 and IoU scores of the CLB decreased by 7.0\% and 6.1\%, respectively. Such results verify the superiority of the feedback mechanism in enhancing the intermediate features for improving forgery localization performance.

{\textbf{Effectiveness of CSPM.}} To investigate the effectiveness of tailored components, we compare the performance of the localization networks with and without the CSPM module. Owing to the CSPM, 16.7\%, 3.6\%, 7.4\%, and 8.8\% increases of the IoU value are achieved on the Columbia, CASIAv1, DSO and AutoSplice datasets, respectively. Furthermore, we experiment with different variants of CSPM, including the CoT block alone and the CSPM using traditional max pooling without dilated convolutions. The test results show that F1 and IoU of the scheme with dilated convolutions are higher than those without dilated convolutions by an average of 6.2\% and 5.7\%, respectively. These results indicate that the CSPM module incorporating dilated convolutions can effectively enhance localization performance by enlarging the receptive field without reducing feature map resolution.

\begin{table*}
\caption{Robustness comparison of different forgery localization algorithms against post-processing of different online social networks (OSNs). Metric values are in percentage. \label{tab:table8}}
\centering
\footnotesize
\begin{tabular}{lccc|cc|cc|cc|cc|cc}
\toprule
\multirow{2.5}{*}{Method} & \multirow{2.5}{*}{OSNs}     & \multicolumn{2}{c}{Columbia} & \multicolumn{2}{c}{CASIAv1} & \multicolumn{2}{c}{NIST16} & \multicolumn{2}{c}{DSO} & \multicolumn{2}{c}{Average}  & \multicolumn{2}{c}{Decline} \\ \cmidrule{3-14}
&   & F1 & IoU   & F1   & IoU   & F1  & IoU   & F1  & IoU   & F1  & IoU    & F1  & IoU  \\ \midrule
PSCC \cite{liu2022pscc}     & \multirow{5}{*}{Wechat}   & 77.5  & 66.2  & 31.9    & 26.9  & 1.7  & 1.0   & 4.5  & 2.6  & 28.9   & 24.2 &31.0 &28.0    \\
OSN \cite{wu2022robust}   &   & 72.7    & 63.1   & 40.5    & 35.8   & 28.6   & 21.4  & 36.6    & 25.2      & 44.6   & 36.4  &10.7 &11.7     \\
CAT \cite{kwon2022learning}    &    & 84.8 & 80.8    & 13.9   & 10.6         & 19.1   & 14.9 & 1.7    & 1.0     & 29.9      & 26.8   &47.7 &47.1  \\
TruFor \cite{guillaro2023trufor} & &77.0 &69.8 &57.3 &51.1 &34.1 &26.1 &44.6 &32.3 &53.2 &44.8  &26.0 &31.9 \\
ProFact    &    & 75.5   & 72.9  & 50.5   & 44.1    & 38.5   & 32.2       & 39.9     & 30.4   & 51.1    & 44.9 &11.2 &12.5   \\ \midrule
PSCC \cite{liu2022pscc}    &\multirow{5}{*}{Facebook}   & 77.4   & 66.3    & 39.7    & 34.6  & 5.2   & 3.2  & 0.9    & 0.6  & 30.8  & 26.2  &26.5 &22.1   \\
OSN \cite{wu2022robust}   &     & 71.4   & 61.1   & 46.2   & 41.7  & 32.9  & 25.3   & 44.7 & 32.0   & 48.8  & 40.0  &2.3 &2.9  \\
CAT \cite{kwon2022learning}   &   & 91.8  & 90.0   & 63.3   & 55.9  & 15.1   & 11.9  & 12.1  & 9.8   & 45.6   & 41.9  &20.2 &17.4  \\
TruFor \cite{guillaro2023trufor}  &   & 74.9 & 67.1 & 67.2 & 60.5 & 33.3 &26.3 &67.4 &57.2 &60.7 &52.8 &15.6 &19.9 \\
ProFact  &  & 74.9    & 72.4    & 55.0   & 48.6    & 42.1   & 35.4    & 45.7    & 35.6    & 54.4    & 48.0   &5.5 &6.4 \\ \midrule
PSCC \cite{liu2022pscc}    & \multirow{5}{*}{Whatsapp}    & 77.0   & 65.7    & 40.4   & 35.2     & 5.7         & 3.6    & 1.1     & 0.6   & 31.0    & 26.3  &25.9 &21.8    \\
OSN \cite{wu2022robust}   &    & 72.7    & 62.8   & 47.8   & 43.1   & 31.3  & 23.9       & 34.1      & 23.3    & 46.5    & 38.3  &7.0 & 7.1     \\
CAT \cite{kwon2022learning}     &   & 92.1   & 89.9     & 42.3     & 37.8         & 20.1    & 16.8    & 2.2     & 1.5    & 39.2     & 36.5   &31.4 &27.9   \\
TruFor \cite{guillaro2023trufor} & &74.8 &66.7 &66.3 &59.9 &38.3 &30.2 &38.8 &30.0 &54.6 &46.7 &24.1 &29.1 \\
ProFact    &      & 76.9     & 74.3  & 55.3   & 49.0    & 39.7     & 33.9    & 41.6  & 31.6  & 53.4   & 47.2  &7.3 &8.0  \\  \midrule
PSCC \cite{liu2022pscc}   & \multirow{5}{*}{Weibo}   & 76.0      & 65.2   & 38.7    & 32.7   & 1.1    & 0.8    & 0.1    & 0.0   & 29.0    & 24.7  &30.9 &26.6   \\
OSN \cite{wu2022robust}  &   & 72.4  & 62.6    & 46.6       & 42.1    & 29.4    & 21.9   & 37.0   & 25.3   & 46.4    & 38.0  &7.2 &7.8    \\
CAT \cite{kwon2022learning}    &    & 92.1   & 89.7     & 42.5    & 36.2     & 20.8    & 16.0  & 2.3     & 1.3     & 39.4     & 35.8    &31.0 &29.3  \\
TruFor \cite{guillaro2023trufor} & &79.2 &72.2 &63.9 &57.7 &32.1 &25.2 &47.2 &36.6 &55.6 &47.9 &22.7 &27.2 \\
ProFact    &   & 75.1   & 72.5   & 52.6      & 46.4   & 40.8        & 34.5   & 47.8    &37.0   & 54.1     & 47.6  &6.1 &7.2    \\ \bottomrule
\end{tabular}
\end{table*}

{\textbf{Influence of training dataset.}} The network trained on the samples of our MBH method ($D_{\rm{MBH-COCO}}$) markedly outperforms that on the simple synthetic samples ($D_{\rm{Simple-COCO}}$). The average F1 and IoU on all test datasets achieve 19.5\% and 16.4\% increments due to the application of MBH, respectively. Specifically, for example, the IoU increases by 28.2\%, 22.0\% and 33.7\% on the CASIAv1, DSO and AutoSplice datasets, respectively. Columbia dataset is an early public dataset with tampered images that are noticeably unrealistic and obvious visual artifacts. Such a dataset has a rather similar distribution to the $D_{\rm{Simple-COCO}}$ due to the same generation mode, i.e., the direct splicing. As such, the network trained on $D_{\rm{Simple-COCO}}$ behaves surprisingly well on Columbia (IoU=90.3\%), but degrades to the worst on the other realistic datasets, for example, IoU=12.7\% on DSO. Such results reveal that the training on large-scale simple splicing samples may easily introduce some biases, which lead to excellent performance on aligned datasets but poor performance on more realistic and distinct samples, such as the AutoSplice dataset (IoU=8.1\%). In contrast, our proposed training sample generation strategy can better simulate the visual realism of real-world forged images, thus establishing the generalization ability of forensic networks.

{\textbf{Influence of two-stage training protocol.}} Table \ref{tab:table7} shows that the performance for the two-stage training surpasses that of the single stage, for example, with an average increment 5.1\% of IoU on all datasets. Since the two-stage strategy reduces the network sensitivity to image scale variations, better generalization performance can generally be gained compared with single-stage training. Note that there exists a slight decrease on CASIAv1 dataset. It should be attributed to the rather low resolution ($384 \times 256$) of CASIAv1 images. Although the second stage of training sacrifices some performance on particularly low-resolution images, it still achieves competitive results, i.e., with F1=56.4\% and IoU=50.2\%.

\begin{figure*}[!tb]
\centering
\includegraphics[width=\linewidth]{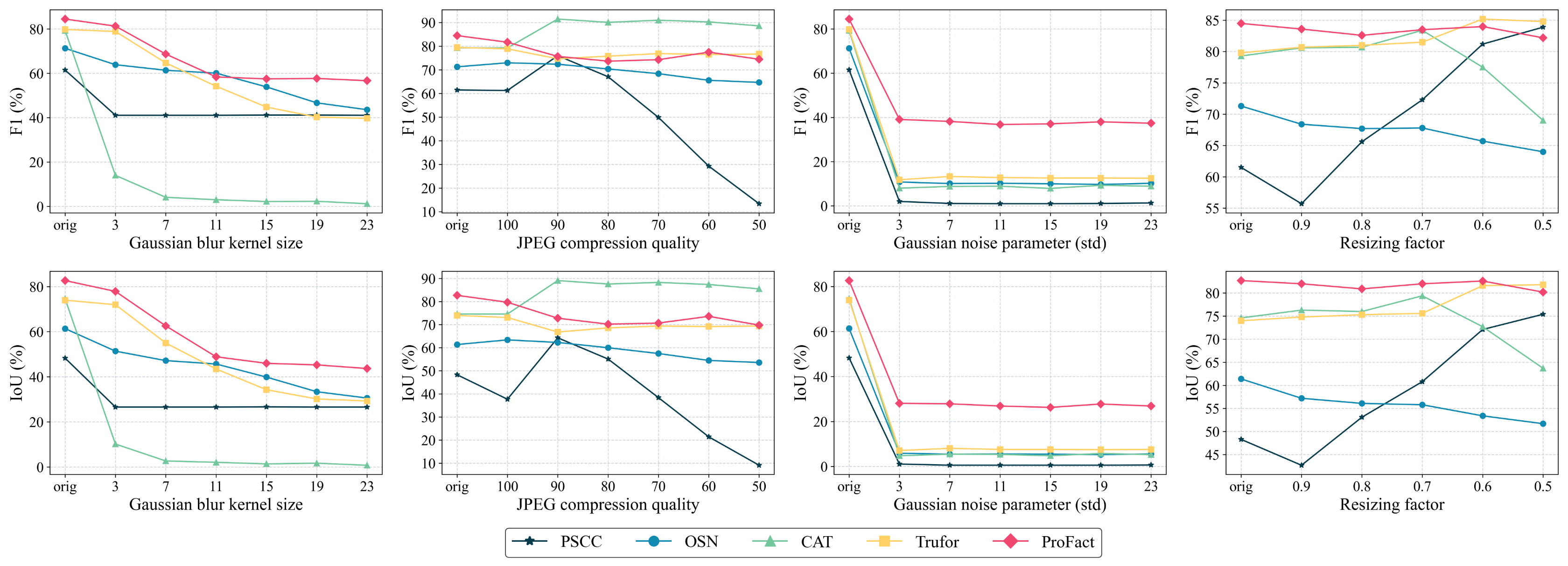}
% \captionsetup{skip=-5pt}
\vspace{-0.5cm}
\caption{Robustness evaluation results of different localization methods against different post-processing operations on Columbia dataset.} \label{fig_7}
\end{figure*}

\subsection{Robustness Analysis}
We first assess the robustness of localizers against the complex post operations introduced by online social networks. Following the prior work \cite{wu2022robust}, four forgery datasets transmitted through Facebook, Weibo, WeChat and Whatsapp platforms are used to evaluate the top-ranking localization methods in the baseline test of Subsection IV-B. As shown in Table \ref{tab:table8}, our method consistently achieves comparable average accuracy across the four datasets for each social network platform, with F1 scores are all above 51\%. Among the compared localizers, the postprocessing results in only a minor performance decline for ProFact. It should be acknowledged that the OSN scheme \cite{wu2022robust} is the pioneer work against social network postprocessing, and also behaves well. However, our method can still exceed the OSN to some extent, for instance, with 8.5\%, 8.0\%, 8.9\%, and 9.4\% increments of IoU on the four datasets, respectively. Note that CAT achieves higher performance on the processed Columbia dataset due to its specialized learning of JPEG compression artifacts. However, it shows an obvious performance decline on the other primarily compressed format of datasets, for instance, with IoU=10.6\% on the Wechat version of CASIAv1. Such results prove the robustness of ProFact against the real social network post-processing.

\begin{table}[!tb]
\caption{Robustness of different localizers against different qualities of post JPEG compression on autoslice dataset. Metric values are in percentage. \label{tab:table9}}
\centering
\begin{tabular}{ccc|cc|cc}
\toprule
\multirow{2.5}{*}{Method} & \multicolumn{2}{c}{Q=75} & \multicolumn{2}{c}{Q=90} & \multicolumn{2}{c}{Q=100} \\ \cmidrule{2-7}
 & F1   & IoU   & F1   & IoU    & F1     & IoU        \\ \midrule
PSCC   \cite{liu2022pscc}     & 3.0     & 1.8    & 6.5    & 4.2   & 60.4    & 49.0              \\
OSN \cite{wu2022robust}   & 24.0    & 18.0  & 35.8      & 26.9             & 50.9    & 39.5              \\
CAT \cite{kwon2022learning}   & 40.2    & 31.8     & 78.9     & 69.9     & 86.2      & 78.0              \\
TruFor \cite{guillaro2023trufor} &29.7 &21.2 &43.4 &32.9 &65.7 &54.7 \\
ProFact (ours)     & 52.3    & 41.9   & 54.7    & 44.0   & 65.5    & 54.4  \\ \bottomrule
\end{tabular}
\end{table}

\begin{figure}[!b]
\vspace{-0.5cm}
\centering
\includegraphics[width=\linewidth]{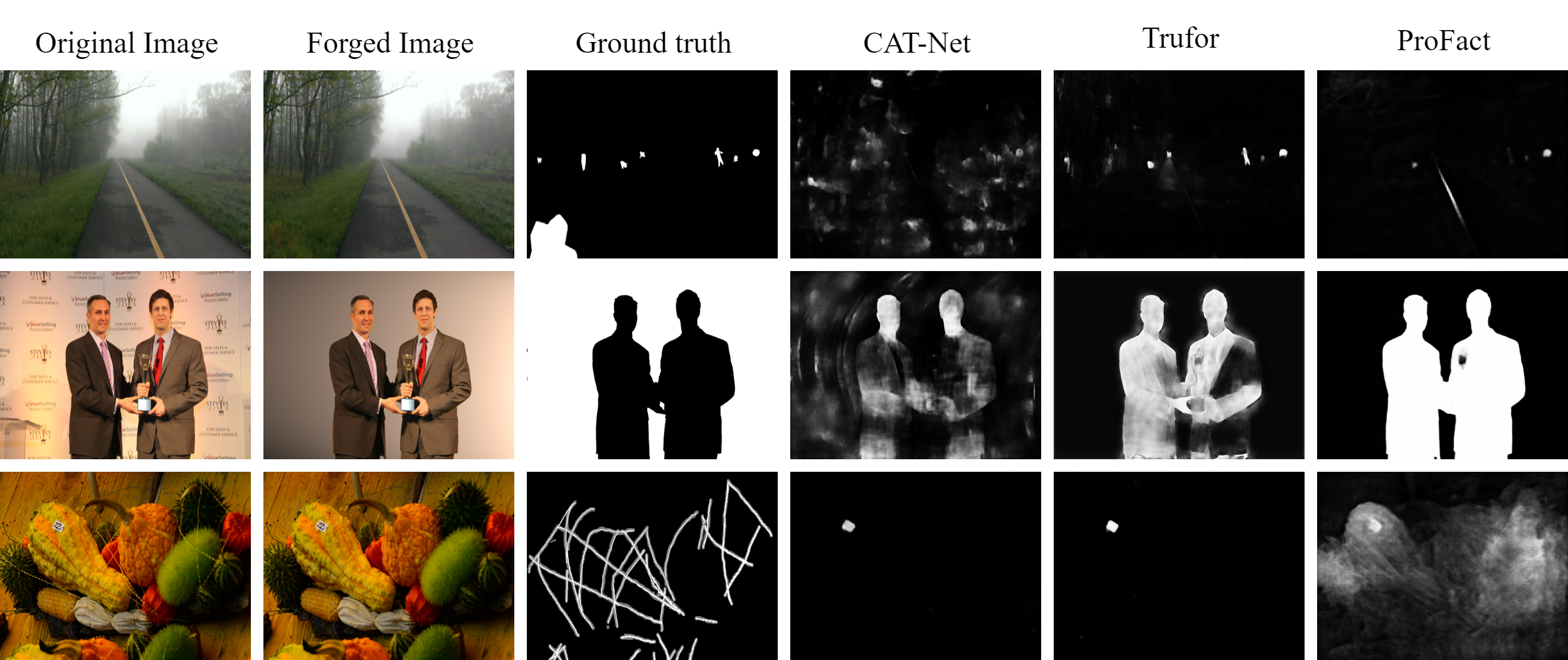}
\caption{Failure cases. Data source: IMD, NIST16 and Korus.}
\label{fig_8}
\end{figure}

Table \ref{tab:table9} shows the localization results for the public AutoSplice datasets at different compression qualities. Our scheme achieves the highest accuracy IoU=41.9\% for the quality factor Q=75, and the second-best IoU=44.0\% for Q=90. These results exhibit the strong robustness of ProFact with respect to post compression in detecting the local AI synthetic regions. Following the prior works \cite{dong2022mvss,liu2022pscc}, the robustness against the post JPEG compression, Gaussian blur, Gaussian noise and resizing is evaluated on the Columbia dataset. The results shown in Fig. \ref{fig_7} also verify the high robustness of our scheme against such post manipulations. The ProFact always behaves the best except for the compression in which the specially tailored network CAT performs better. Although the noising heavily degrades the performance of all localizers, the ProFact can still achieve acceptable accuracy, i.e., about IoU=30\%, which is much higher than that of other localization methods (below IoU=15\%). Gaussian noise, characterized by its random distribution and varying intensity levels, distorts forensic features and hinders classifiers from discerning genuine from forged regions. The high robustness of ProFact demonstrated by such results should be attributed to the applied data augmentation and the two-stage training protocol.

\subsection{Limitations}
Given the challenging nature of the image forgery localization task, the performance limitations of our proposed scheme are discussed by detailedly analyzing some failure cases. As shown in Fig. \ref{fig_8}, the forged regions of the first example image blend seamlessly into the foggy background. In addition, the whole image has undergone severe blurring. We can see that the CAT, TruFor and our ProFact methods all fail to discover the major forgery region at the bottom left. For the second example, although all the methods can locate the inconsistent region, they have not realized that the background is actually falsified. The last example shows the failures in detecting the removal type of forged images, in which the inpainting region closely resembles the background or is too small, such as the inconspicuous lines. As for the state-of-the-art localization models, the available forensic traces behind such hard forgery samples in extreme cases appear too weak to be used for successfully identifying the forged regions. Solving such problems should be the direction of efforts in future work towards real-world forensic applications.

\section{CONCLUSION}
In this paper, we propose a novel progressive feedback-enhanced network for accurate and robust image forgery localization. The network is built on two cascaded transformer branches, which feed the generated coarse localization map back to the early encoding layer for further refinement. Furthermore, an effective image composition strategy is presented to automatically generate plentiful realistic forged samples. Such samples are better aligned with real-world applications, ensuring efficient learning of forensic features to achieve outstanding performance with reduced training costs. A progressive two-stage training protocol is further used to ensure the reliability and robustness of our proposed localization network. Extensive experimental results demonstrate the superior performance of our scheme compared with the state-of-the-art on multiple benchmarks, including the effectiveness against unknown tampering operations. In future work, it is promising to study more powerful forensic algorithms for successfully addressing the hard forgery samples in extreme cases and the rapidly evolving tampering techniques.

% \begin{thebibliography}{1}
\bibliographystyle{IEEEtran}
\bibliography{ref}

\vfill

\end{document}